\def\year{2022}\relax
\documentclass[letterpaper]{article} % DO NOT CHANGE THIS
\usepackage{aaai22}  % DO NOT CHANGE THIS
\usepackage{times}  % DO NOT CHANGE THIS
\usepackage{helvet}  % DO NOT CHANGE THIS
\usepackage{courier}  % DO NOT CHANGE THIS
\usepackage[hyphens]{url}  % DO NOT CHANGE THIS
\usepackage{graphicx} % DO NOT CHANGE THIS
\usepackage{natbib}  % DO NOT CHANGE THIS AND DO NOT ADD ANY OPTIONS TO IT
\usepackage{caption} % DO NOT CHANGE THIS AND DO NOT ADD ANY OPTIONS TO IT
\DeclareCaptionStyle{ruled}{labelfont=normalfont,labelsep=colon,strut=off} % DO NOT CHANGE THIS
\usepackage{algorithm}
\usepackage{algorithmic}
\usepackage{mathtools}
\usepackage{makecell}
\usepackage{bbding}
\usepackage{amssymb}% http://ctan.org/pkg/amssymb
\usepackage{pifont}% http://ctan.org/pkg/pifont
\usepackage{newfloat}
\usepackage{listings}
\floatstyle{ruled}
\newfloat{listing}{tb}{lst}{}
\floatname{listing}{Listing}
\title{Fine-Grained Control of Artistic Styles in Image Generation}
\author{Xin Miao, Huayan Wang, Jun Fu, Jiayi Liu, Shen Wang, Zhenyu Liao}
\affiliations{Kuaishou Technology\\ xinmiao@kuaishou.com, wanghuayan@kuaishou.com, fujun@kuaishou.com, 
             jiayiliu@kuaishou.com,  wangshen@kuaishou.com, liaozhenyu2004@gmail.com}

\begin{document}
\maketitle

\begin{figure*}[t]
\begin{center}
\includegraphics[width=0.85\linewidth]{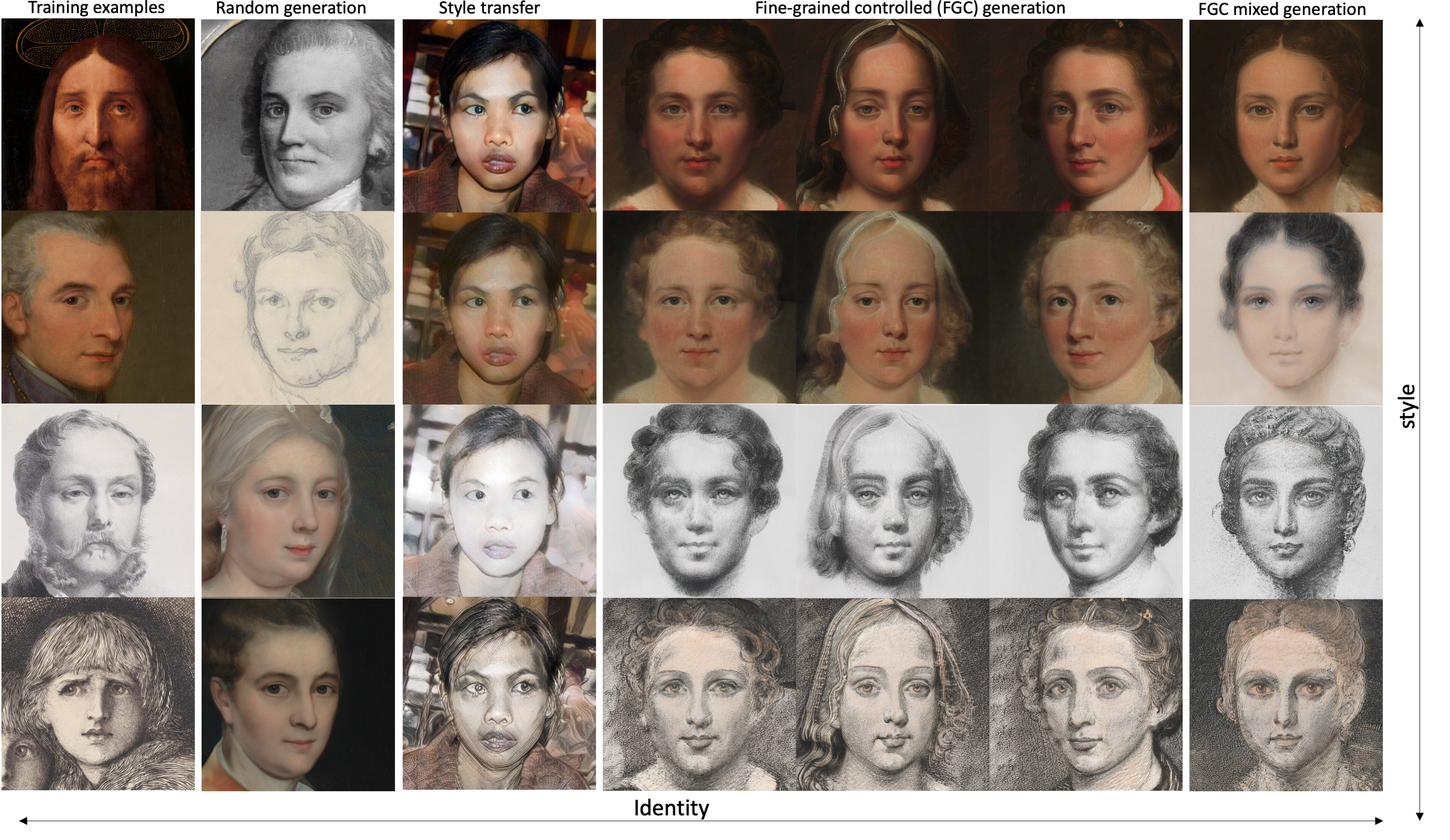}
  
\caption{Overview. {\bf 1st column}: Examples of training images (from 1,336) \cite{karras2020training} of varied artistic style. {\bf 2nd column}: Generated images of mixed style by learning from all training examples \cite{karras2019style}. {\bf 3rd column}: Attempting to perform style control by transferring  \cite{huang2017arbitrary} from each individual example in the 1st column does not provide satisfactory results. {\bf 4th-6th column}: Our results with precise style control. Each column renders the same person in different styles; each row represents the style of the corresponding image in the 1st column. {\bf 7th column:} New artistic styles created by interpolating between the style of that row (as represented by the 1st column images) and the next row, wrapping around at the 4th row.}
\label{fig:intro}
\end{center}
\end{figure*}

\begin{abstract}
Recent advances in generative models and adversarial training have enabled artificially generating artworks in various artistic styles. It is highly desirable to gain more control over the generated style in practice. However, artistic styles are unlike object categories---there are a continuous spectrum of styles distinguished by subtle differences. Few works have been explored to capture the  continuous spectrum of styles and  apply it to a style generation task. In this paper, we propose to achieve this by embedding original artwork examples into a continuous style space. The style vectors are fed to the generator and discriminator to achieve fine-grained control. Our method can be used with common generative adversarial networks (such as StyleGAN). Experiments show that our method not only precisely controls the fine-grained artistic style but also improves image quality over vanilla StyleGAN as measured by FID.
\end{abstract}

\begin{figure*}[t]
\begin{center}
\includegraphics[width=0.95\linewidth]{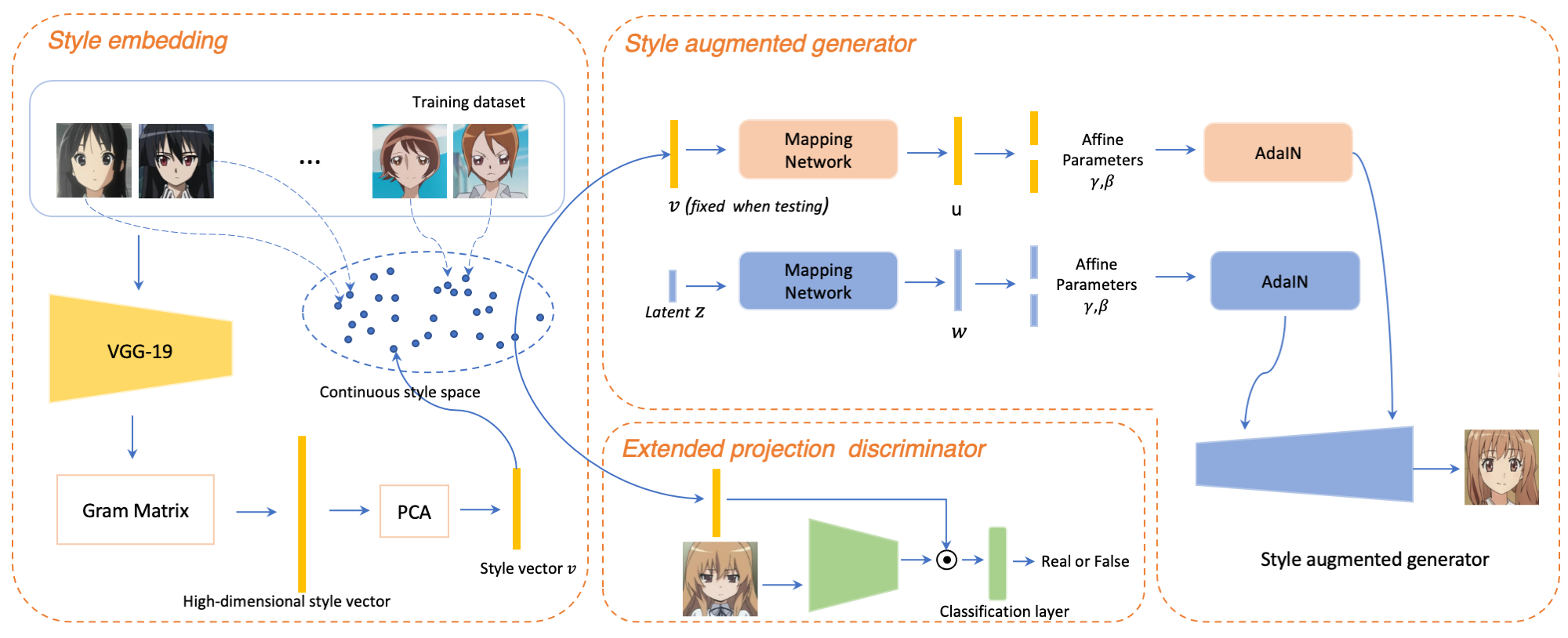}
\end{center}
\vspace{-0.3cm}
\caption{Overview of the proposed framework. First, all collected images are embedded in a continuous space by using Gram matrix \cite{gatys2016image} and PCA. Spatial proximity of the embedded style vectors well reflects the visual style similarity of the  original images. Style vector is sampled and passed through the mapping network. It is used as affine parameters to modulate the image generator's AdaIN modules. A projection discriminator is used to take into account the style vector and supervise the generator. Combining these components we are able to precisely control the style of the generated images to match that of each single training example or to interpolate between them using their corresponding style vectors. }
\label{fig:long}
\label{fig:onecol}
\end{figure*}

\section{Introduction}\label{sec:intro}

% intro to image generation
Image generation in various forms has been a long quest in the vision community, not only for its practical use and aesthetic value, but also for assessing our model's ability to disentangle and exploit different aspects of image variability. Separating content and style in image generation dates back to the seminar work of Tenenbaum and Freeman \cite{nips-9:Tenenbaum+Freeman:1997} more than twenty years ago. Since then, resorting to a hierarchical architecture has enabled us to generate images of more structured objects such as human faces \cite{conf/icml/LeeGRN09,journals/corr/KingmaW13}. More recently, adversarial training \cite{NIPS2014_5ca3e9b1} and improved model architecture design \cite{karras2020analyzing} have brought the quality of computer-generated images to a level that is oftentimes indistinguishable from real images to our eyes. Besides generating photo-realistic images, we can also generate images of various artistic styles by learning from artwork examples \cite{gatys2016image, zhu2017unpaired, liu2017unsupervised}.

%The latter is one of the fundamental questions of computer vision. Early work in texture modeling and synthesis \cite{bb26978}\cite{journals/ijcv/ZhuGWX05} was deeply rooted in the attempt to understand how natural images are formed from a statistical point of view. 

Our goal in this paper is to gain fine-grained control of the artistic style in generating human portrays. Compared to generating other types of artwork, human portray generation poses a greater challenge in image quality, because our perception system is highly sensitive to human face details. Precisely controlling the artistic style of generated images is often desirable in practical applications. For example, in the data preparation stage of distilling highly efficient mobile-friendly models for style transfer, it is essential for the result images to have very uniform artistic style to ensure the stability of subsequent model distilling steps.

The target artistic style is usually given by artwork examples. However, the uniformity of style is extremely difficult to guarantee in real applications---even the same artist would often show subtle style variations in different works. Moreover, as we try to collect more data---which we all know would tremendously benefit most learning algorithm---we inevitably bring in data of slightly different artistic styles (\emph{e.g.}, Fig.~\ref{fig:intro} 1st column). As a result, learning without accounting for their differences would result in a model representing a mixture of artistic styles that we have no control over (\emph{e.g.},  Fig.~\ref{fig:intro} 2nd column). 

On the other hand, one may attempt to ensure the exact artistic style by learning from a single example. Because of limited information, we have to settle for a low-capacity representation of style~\cite{gatys2016image, huang2017arbitrary}. As a result, we can only generate images of rough artistic styles that are rather unconstrained in details. For example, style transfer \cite{huang2017arbitrary} from each individual example ( Fig.~\ref{fig:intro} 3rd column) cannot achieve satisfactory visual quality.

Therefore it becomes a key factor whether our algorithm can effectively make use of many examples that are too relevant to abandon but also has non-negligible differences. To tackle this, we introduce a framework that effectively learns from many examples of slightly varying artistic style while providing fine-grained control over the style of the generated images. Our method consists of three major components: 1) a continuous embedding space of training examples; 2) a style augmented generator architecture; and 3) a projection discriminator \cite{miyato2018cgans} extended to account for the style embedding. Combining these components we are able to precisely control the style of generated images to match that of each single training example (\emph{e.g.}, Fig.~\ref{fig:intro} 4th to 6th column) or to interpolate between them (\emph{e.g.}, Fig.~\ref{fig:intro} 7th column).

Our framework is general in the sense that the components can be used with different existing image generation or style transfer methods. In particular, we show how to use it with StyleGAN \cite{karras2019style, karras2020analyzing} and CycleGAN \cite{zhu2017unpaired}, which are representative of modern methods in these categories. The results for fine-grained controlled CycleGAN can be found in the supplementary materials.

The effectiveness of the proposed fine-grained control has been verified by extensive experiments for image synthesis. The controlled version StyleGAN can provide fine-grained control over the style of the generated images. It is evaluated on three datasets including MetFaces, Selfie2anime and our own game-figure dataset with a very small number of (78) training images. The small-sample case emphasizes our approach's ability to leverage limited data and generalize well. Compared with the original StyleGAN, our framework can improve the quality for the generated images (as measure by FID) which largely surpass the recent GAN-control work \cite{shoshan2021gan}. 

Our main contributions are summarized in the following  aspects.
 
\begin{itemize}
  \item [1)] 
  We propose to embed original artwork examples into a continuous style space by using the Gram matrix of extracted feature vectors. The space captures the continuously varying artistic style of images. Using the Gram matrix-based embedding also makes our approach \emph{unsupervised}, which gives it great potential to exploit a larger set of artworks on the Internet.
  
  \item [2)]
  We propose a style augmented generator architecture that incorporates the style vector into the model. Together with the extended projection discriminator, our model can be trained by accounting for the fine-grained artistic style differences within the training data. As a result we can offer fine-grained control of artistic styles in image generation by providing a specific style vector at test time.
  
  \item [3)]
  To the best of our knowledge, we first address the  fine-grained artistic style control problem. And our approach can be easily injected to state-of-art GANs to provide fine-grained artistic style control.  We incorporate our framework into StyleGAN and CycleGAN (in supplementary materials), and verify them on two public datasets as well as our own game-figure dataset. Our method offers much better results than the state-of-the-art \cite{shoshan2021gan} for pure image generation. 

\end{itemize}

%------------------------------------------------------------------------
\section{Related Work}

\textbf{Generative Adversarial Networks.} Generative Adversarial Networks \cite{NIPS2014_5ca3e9b1} have achieved impressive performance in lots of computer vision tasks \cite{zhu2017unpaired, miyato2018spectral, isola2017image, choi2018stargan, brock2018large, choi2020stargan, chen2018cartoongan, wang2018high}. StyleGAN \cite{karras2019style} proposed an alternative generator architecture for generative adversarial networks, borrowing from style transfer literature. StyleGAN and other state-of-the-art works \cite{brock2018large, karras2017progressive} can generate high quality synthetic images which are almost indistinguishable from real images.  \cite{karras2020training} significantly stabilizes training in limited data regimes by an adaptive discriminator augmentation mechanism. \cite{karras2020training} also proposes an artistic styles dataset and they show good image generation result on it by only using limited training data.  But the learning process doesn't account for the differences between training examples which can't provide fine-grained style control. Unlike the unconditional (original) GAN, conditional GANs use conditional information for the generator and discriminator. GAN-based techniques have also been widely explored in the image-to-image translation task. For supervised training, Pix2Pix \cite{isola2017image} uses L1 loss with a cGAN framework to learn the mapping network from source domain images to target domain images. For unsupervised training, in order to get rid of the dependency on paired data, a cycle-consistency loss was proposed by CycleGAN \cite{zhu2017unpaired}. We show our framework is general in the sense that the components can be used in either image synthesis framework (StyleGAN) or image-to-image translation framework (CycleGAN). Unlike the unconditional (original) GAN, conditional GANs use conditional information for the generator and discriminator. Instead of naively concatenating conditional information to the input and feeding them to discriminators, \cite{miyato2018cgans} proposed a novel, projection based method to incorporate the conditional information into GAN discriminators.

\textbf{Control for image generation.} Many approaches \cite{jahanian2019steerability,yang2021semantic,shen2020interpreting,balakrishnan2020towards, harkonen2020ganspace} exploit the inherent disentanglement properties of GAN latent space to control the generated images. InfoGAN~\cite{chen2016infogan} maximizes the mutual information between the observation and a subset of the latent variables. InfoGAN successfully disentangles writing styles from digit shapes on the MNIST dataset in a completely unsupervised manner. Some methods \cite{deng2020disentangled, kowalski2020config, tewari2020stylerig} were proposed to control face image generation explicitly. However, they are only applicable to the controls parameterized by a 3D face model. \cite{shoshan2021gan} uses contrastive learning to obtain GANs with an explicitly disentangled latent space. In the face domain, they can control over pose, age, identity, \emph{etc.}. In the painted portrait domain, they can also control the style. But similar to \cite{deng2020disentangled, kowalski2020config}, there is a deterioration in FID when control is introduced \cite{shoshan2021gan}. Compared with these methods, our method improves the quality of the generated images over the original non-control version without the need of 3D face rending framework and we embed original artwork examples into a continuous so we can interpolate between them (moving in the style space) at test time. There are some works designed for neural style transfer. \cite{huang2017arbitrary} aligns the mean and variance of the content features with those of the style features by adaptive instance normalization (AdaIN). They use an encoder to embed all style images in a continuous space and can achieve style transfer by interpolating between arbitrary styles. However this kind of methods can't generate high quality facial images by transferring the style from training artistic styles examples, because it's more likely to transfer the texture styles to the target image. Compared with it, our method incorporates the projection discriminator to make the generator output images which close to the distribution of real style images and provide the fine-grained style control.

%------------------------------------------------------------------------
\section{Proposed approach}

\noindent {\bf Overview.} We start by creating an embedding of the training images that reflects their continuously varying artistic style. Vectors in the style embedding space (that we call \emph{style vectors}) are subsequently used in the image generator  and the extended projection discriminator \cite{miyato2018cgans}  that we use to help training. To be more specific we introduce these components in the context of StyleGAN \cite{karras2019style, karras2020analyzing}. Our modules can be used with image translation frameworks as well such as CycleGAN (see supplementary materials).

%-------------------------------------------------------------------------
\subsection{Style embedding}
\label{sec:approach:embedding}
We hope to embed original artwork examples into a continuous style space which reflects their slightly varying artistic style. The Gram matrix has been widely used to extract style information from images~\cite{gatys2015texture,gatys2016image},  it consists of the correlations between different filter responses. We use VGG-19 CNN architecture pre-trained for the ImageNet~\cite{bb77703} classification task to extract Gram matrix features as described in the following. With an artwork image as input, we compute the the activations for each convolution layer of the network. 

\begin{figure}[t]
\begin{center}
\includegraphics[width=\linewidth]{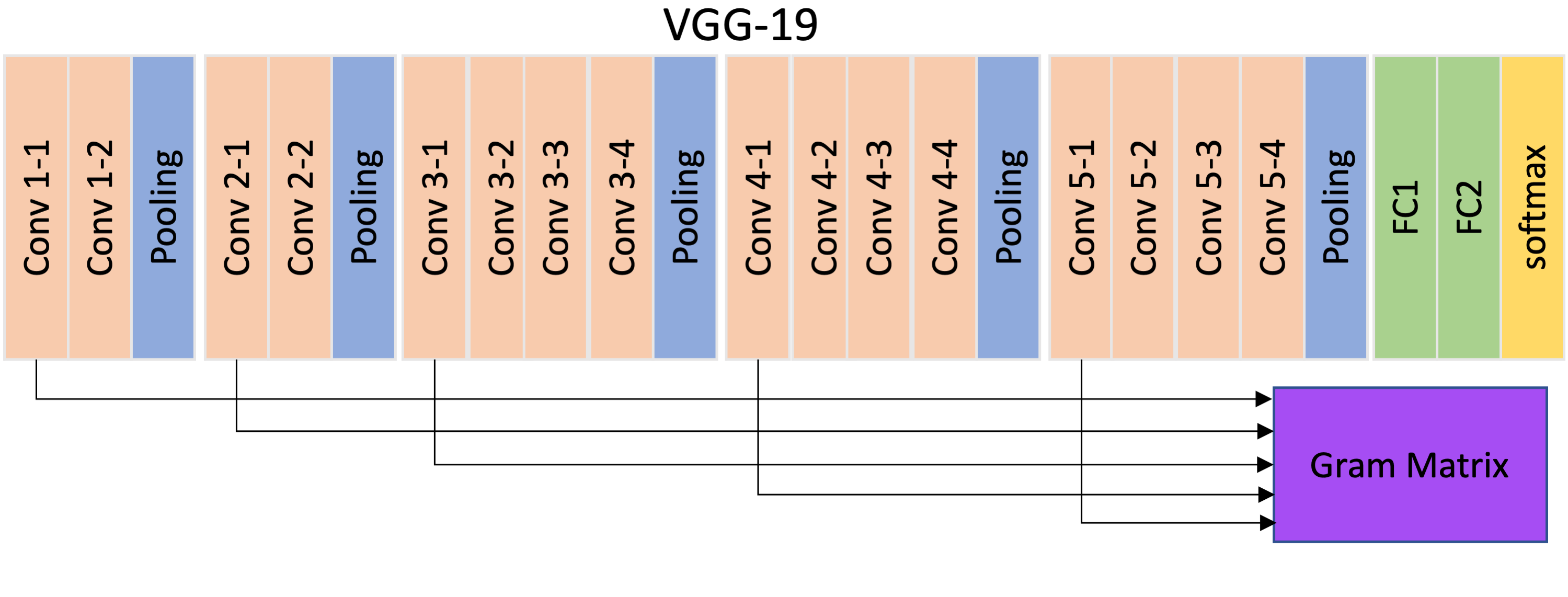}  
\end{center}
\vspace{-0.4cm}
\caption{Gram matrix is calculated by the features from 5 layers of VGG-19 network.}
\label{fig:vgg}
\end{figure}

Consider $N$ feature maps corresponding to $N$ distinct filters in a convolution layer $l$, each with size $M_{l} = H_{l} \times W_{l}$, where $H_{l}$ and $W_{l}$ are the height and width of the feature map in layer $l$, respectively. We use a matrix $F^{l}$ to store the activations in layer $l$, where $F_{i,x,y}^{l}$ is the activation of the $i^{th}$ filter at position $(x,y)$ in layer $l$.  Gram matrix is defined as:
\begin{equation}
G_{i,j}^{l} =\sum_{x,y} F_{i,x,y}^{l}F_{j,x,y}^{l}
\end{equation}
In order to obtain a multi-scale representation for the input image, we include the Gram matrices on five layers of VGG-19 network, namely \emph{conv1-1}, \emph{conv2-1}, \emph{conv3-1}, \emph{conv4-1}, \emph{conv5-1} as shown in Fig.~\ref{fig:vgg}.

% figure

The resulted Gram matrices are flattened and concatenated into a very high dimensional ($\sim$ 610,000) vector. In order to make the computation more efficient, Principal Component Analysis (PCA) is applied to reduce its dimensionality to 72 or 512 (depending on the size of the dataset) and it preserves at least 99\% of the total variation. We call the extracted  vector the \emph{style vector}.

\begin{figure}[t]
\begin{center}
\includegraphics[width=\linewidth]{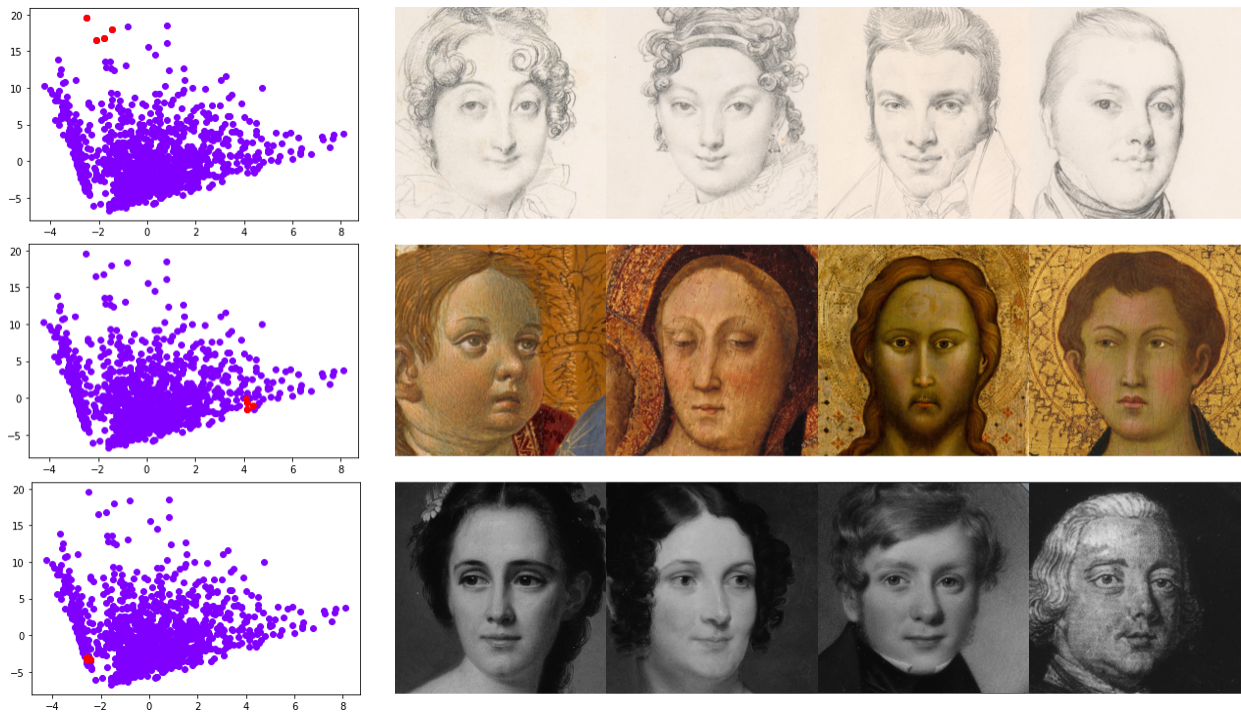}  
\end{center}
\vspace{-0.2cm}
\caption{Visualization of style vectors on MetFaces by using PCA (left). The style vectors corresponding to the right images are those red points in the left plot.}
\label{fig:figure4}
\end{figure}

For visualization purpose we further reduce the dimensionality of the style vectors to two using PCA. A typical embedding is shown in Fig.~\ref{fig:figure4}. We can see that spatial proximity of the embedded style vectors well reflects the visual style similarity of the original images. Note that the 2D embedding is for visualization only. In the generator and discriminator (subsequent sub-sections) we use the 72-D or 512-D style vectors.
We can  precisely control the style of generated images to match that of each single training example (manually select style vector) or to interpolate between them (moving in the style space) at test time.

% figure

\subsection{StyleGAN}
\label{sec:approach:bg}
The embedded style vectors will be used to inform an image generator $G$ and discriminator $D$. Before getting into their construction we first put them in context. 
In a typical image generation framework (e.g., StyleGAN~\cite{karras2019style, karras2020analyzing}), the learning objective is:

\begin{equation} \label{Eq:loss_gan}
\begin{split}
\mathcal{L}_{{\rm styleGAN}}(G,D) &= \mathbb{E}_{\mathbf{y}}[\log D(\mathbf{y})]\\
&~~+ 
\mathbb{E}_{\mathbf{z}}[\log (1-D(G(\mathbf{z})))],
\end{split}
\end{equation}
where $G$ and $D$ denotes the generator and discriminator respectively.  The image $\mathbf{y}$ is drawn from the target image distribution. The input vector $\mathbf{z}$ is drawn from a standard normal distribution. In this paper we extend $G$ and $D$ to take an extra style vector input $\mathbf{v}$, \emph{i.e.}~$G(\mathbf{z}, \mathbf{v})$ and $D(\cdot, \mathbf{v})$.

% In image translation (e.g., CycleGAN~\cite{zhu2017unpaired}) from domain A to domain B, the learning objective for the target domain B is:
% \begin{equation} \label{Eq:loss_gan}
% \begin{split}
% \mathcal{L}_{{\rm cycleGAN}}(G_{A\rightarrow B},D_{B}) & = \mathbb{E}_{\mathbf{y}}[\log D_{B}(\mathbf{y})]\\
% &~~+ 
% \mathbb{E}_{\mathbf{x}_{A}}[\log (1-D_{B}(G_{A\rightarrow B}(\mathbf{x}_{A})))],
% \end{split}
% \end{equation}
% where the image $\mathbf{y}$ is drawn from target domain B; $\mathbf{x}_{A})$ is drawn from source domain A; $D_B$ is the discriminator in domain B; $G_{A\rightarrow B}$ is image translator. In this paper we assume that domain B is where we have continuously varying artistic style and the style vector embedding. We extend the $G_{A\rightarrow B}$ and $D_B$ to take an extra style vector input $\mathbf{v}$, \emph{i.e.}~$G_{A\rightarrow B}(\mathbf{x}_{A}, \mathbf{v})$, $D_B(\cdot, \mathbf{v})$. The generator in the other direction $G_{B\rightarrow A}(\mathbf{x}_{B})$ and the corresponding learning objective function remain the same.

%-------------------------------------------------------------------------
\subsection{Style vector modulated image generator}
\label{sec:approach:generator}

One of the basic building block of a generator $G$ is the adaptive instance normalization (AdaIN) module \cite{huang2017arbitrary} defined as:
\begin{equation}
\mathbf{AdaIN}(\mathbf{f}, \gamma, \beta) = \gamma(\frac{\mathbf{f}-\mu(\mathbf{f})}{\sigma(\mathbf{f})})+\beta
\end{equation}
which first normalizes the input feature map $\mathbf{f}$, then applies a learnable transform $(\gamma, \beta)$ to it.

The AdaIN module is widely used in many generative architectures including StyleGAN~\cite{karras2020analyzing}. Our generator extends \cite{karras2020analyzing} with the style vector incorporated to modulate the generation process. Specifically, StyleGAN~\cite{karras2020analyzing} uses an eight-layer nonlinear mapping network to map from a Gaussian-distributed vector $\mathbf{z}$ to $\mathbf{w}$, which is subsequently used to construct the AdaIN transform parameters. So the generator can control different attributes at each convolution layer based on the latent code. The intermediate latent space is allowed to be disentangled. We add another eight-layer nonlinear mapping network that maps our style vector $\mathbf{v}$ to $\mathbf{u}$. Then we use $\mathbf{u}$ and $\mathbf{w}$ together ($\mathbf{u}$ for last 4 feature maps that are close to output, $\mathbf{w}$ for other feature maps) to construct the AdaIN modules of StyleGAN.  In StyleGAN, varying $\mathbf{z}$ changes various attributes (\emph{i.e.}~face identity) of the generated image. In our extended generator, the style vector control the style attributes (\emph{i.e.}~ texture, color), so varying the style vector $\mathbf{v}$ (therefore $\mathbf{u}$) while keeping $\mathbf{z}$ (therefore $\mathbf{w}$) fixed would generate different artistic styles of the same person. Similarly, varying $\mathbf{z}$ with a fixed $\mathbf{v}$ would generate different images of the same artistic style. Using the Gram matrix-based embedding also makes our approach unsupervised, which gives it great potential to exploit a larger set of artworks on the Internet.

%Our aim is to generate images with controllable fine-grained style. We take the previous extracted style vector as input to the control block of the generator. Inspired by~\cite{karras2020analyzing}, an eight-layer nonlinear mapping network is first devised to encode the affine transformations, which can benefit the generator manipulated fine-grained information. Adaptive instance normalization (AdaIN) embeds the transformed vector to transform the feature. The AdaIN is 
%where each  feature $x$ is normalized independently. $\gamma$ and $\beta$ are the scale and bias vectors learned from the fine-grained style label. 

Details on how to adapt this generator design to StyleGAN~\cite{karras2020analyzing} will be given in the experiment section.

\subsection{Extended projection discriminator with the style vector}
\label{sec:approach:discriminator}
In order to generate the images condition on the style vector, we need to feed the style vector $\mathbf{v}$ to the generator and discriminator during training. A projection based discriminator to incorporate the conditional information into GANs was proposed in~\cite{miyato2018cgans}.  In contrast with most methods using the conditional information by concatenating the (embedded) conditional vector to the feature vector or input, the projection discriminator takes an inner product between the embedded condition vector and the feature vector. It was proved to be able to significantly improve the quality of class conditional image generation. In our framework, $D(\cdot, \mathbf{v})$ uses the style vector $\mathbf{v}$ as the conditional information, so the inner product is between the style vector and the feature vector before feeding into the last classification layer of the discriminator.

%Therefore in this paper we extend the discriminator to take an extra style vector input $\mathbf{v}$, i.e.~$D(\mathbf{y}, \mathbf{v}})$ and $D(G(\mathbf{z}), \mathbf{v}})$.

%------------------------------------------------------------------------
\section{Experiments and Results}
In this section, we first describe implementation details and datasets. Then we show the experimental results for style-controlled versions of StyleGAN. We also analyze the style space and conduct ablation study.

\begin{figure}[tp]
\begin{center}
\includegraphics[width=1.0\linewidth]{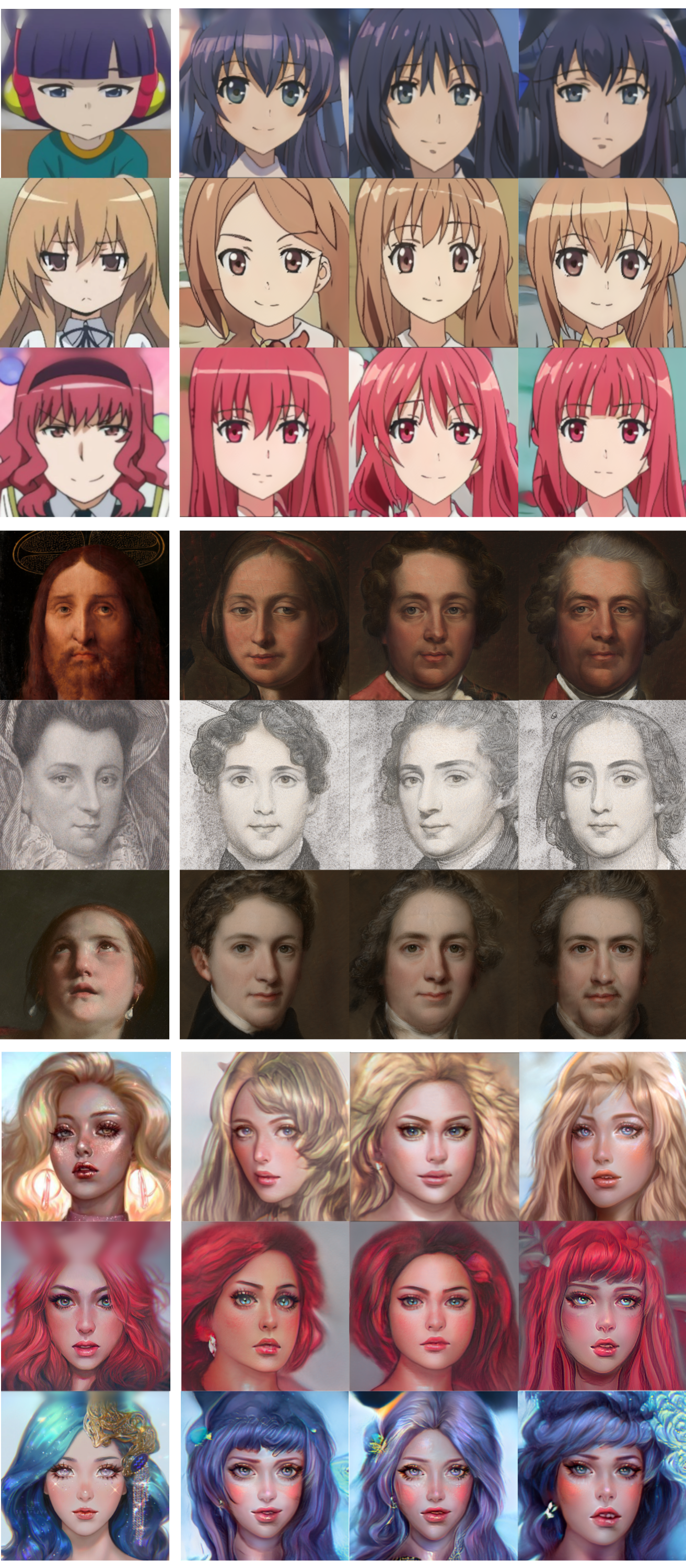}
\end{center}
\vspace{0.2 cm}
\caption{Controlling fine-grained artistic style on different datasets: Selfie2anime (top), Metfaces (middle), Game figure (bottom). The left-most image in each row is a training example; other columns are images generated using its style vector.}
\label{fig:visStyleGAN}
\end{figure}

\subsection{Training}
\label{sec:experiments:Training}
%\textbf{Implementation details.}
%The proposed fine-grained control block can be easily applied to the pure generative model StyleGAN or image translation framework CycleGAN. 

The training setup for StyleGAN in this paper is transfer learning which means the StyleGAN was first trained on FFHQ ~\cite{karras2019style}, then we finetune it in the style domain. Therefore, we reuse the pretrained discriminator and add the projection discriminator to it in the finetuning stage. All images used in our experiments are aligned and cropped by the same preprocessing in StyleGAN \cite{karras2019style}.

\textbf{Datasets.} We conduct our experiments for style-controlled StyleGAN on three artistic datasets: 
\begin{itemize}
\item Metfaces~\cite{karras2020training} with 1,336 high-quality $1024 \times 1024$ artistic faces collected from Metropolitan Museum of Art. 
\item Selfie2anime with 3,400 anime images for training, with an image size of $256 \times 256$. For this dataset we follow the same experimental setting as in UGATIT~\cite{kim2019u}.
\item We also collected our own extremely challenging \emph{game-figure} dataset with only 78 images. All images are collected from the Internet and resized to $512 \times 512$. 
\end{itemize}

\begin{table}
\begin{center}
\begin{tabular}{c|c|c|c}
\hline
Method & metfaces & selfie2anime & game figure \\
\hline
StyleGAN &29.73 &18.62 & 66.57\\
\hline
FGC StyleGAN & {\bf 20.39}& {\bf 14.24} & {\bf 63.64} \\
\hline
\end{tabular}
\end{center}
\caption{FID scores (lower is better) of our proposed fine grained control (FGC) StyleGAN compared with the original StyleGAN on different datasets.}
\label{tab:styleganFID}
\end{table}

\begin{table}
\begin{center}
\begin{tabular}{c|c|c}
\hline
GAN Version & Ours & \cite{shoshan2021gan} \\
\hline
Vanilla & 29.73 &  26.60 \\
\hline
Controlled &  20.39  ($\downarrow$)  & 28.56($\uparrow$)
 \\
\hline
\end{tabular}
\end{center}
\caption{FID scores (lower is better) for different methods on MetFaces. With our method the controlled StyleGAN has a better FID score than the vanilla version. In contrast, GAN-control in \cite{shoshan2021gan} deteriorates the FID score. Due to
different data-preprocessing settings, the FID
scores are not comparable between columns.}

\label{tab:compareWithGANcontrol}
\end{table}

\subsection{Results}
\label{sec:experiments:Results}
Table~\ref{tab:styleganFID} shows FID scores of different methods on three datasets. The FID socre for our controlled version StyleGAN is calculated by random sampling latent codes style vectors. By using our proposed fine-grained control module, StyleGAN always achieves better FID score than the vanilla version. We can see that our method and the Vanilla version have the largest gap on Metfaces. Because Metfaces is an artistic faces collected Metropolitan Museum of Art, slightly different  artistic  styles are inevitably  brought  in  data when collect  artwork examples in real application. And compared with other two datasets, the artwork examples in Metfaces have fine-grained difference not only in texture but also in color. It can prove that our model has the ability to account for the fine-grained artistic style differences within the training data and offer control during testing.

Allowing explicit control over various attributes in StyleGAN generation has been explored in other works. However, fine-grained artistic style control wasn't clearly addressed before. We compare with ~\cite{shoshan2021gan}, which is the most closely related method. They assign discrete labels to images and use contrastive learning to train the model. Our method embed artwork examples into a continuous style space and train our model in unsupervised setting. When control is introduced, previous explicit control methods usually get a deterioration in FID ~\cite{deng2020disentangled, kowalski2020config, shoshan2021gan}. For example, in \cite{shoshan2021gan} the FID increases from 26.60 of the vanilla StyleGAN to 28.56 of the controlled version (Table~\ref{tab:compareWithGANcontrol}). Unlike existing approaches, our method not only provides fine-grained control but also improves FID and image quality. Note that the FID scores are not  directly comparable between different columns in Table~\ref{tab:compareWithGANcontrol}---although they use the same MetFaces dataset---due to different data-preprocessing steps, image resolutions and number of generated images in calculating FID. We are instead emphasizing the different directions of FID changes from the vanilla version to the controlled version in our work and \cite{shoshan2021gan}.

Fig ~\ref{fig:visStyleGAN} shows our generated images with fine-grained controlled StyleGAN. In each row, the left image is a query image selected from the training set. The style vector $\mathbf{v}$ is extracted from that image to control the artistic styles of the output images. The remaining images in each row are generated by our model by fixing the style vector $\mathbf{v}$ and varying the person identity vector $\mathbf{z}$. We can see that the artistic style of the generated images are precisely controlled by the style vector $\mathbf{v}$. Especially on the game-figure dataset, we can achieve high quality image generation with style control given only 78 training images.

\begin{table}[tp]
\begin{center}
\begin{tabular}{c|c}
\hline
Method & FID  \\
\hline
FGC StyleGAN with embedded style space  &  {\bf 20.39}\\
\hline
FGC StyleGAN with random style space&  29.12 \\
\hline
Vanilla StyleGAN&  29.73 \\
\hline
\end{tabular}
\end{center}
\caption{FID scores (lower is better) on Metafaces of FGC StyleGAN with embedded style space versus random style space, compared to vanilla StyleGAN.}
\label{tab:randomlabel}
\end{table}

\begin{figure}[tp]
\begin{center}
\includegraphics[width=1.0\linewidth]{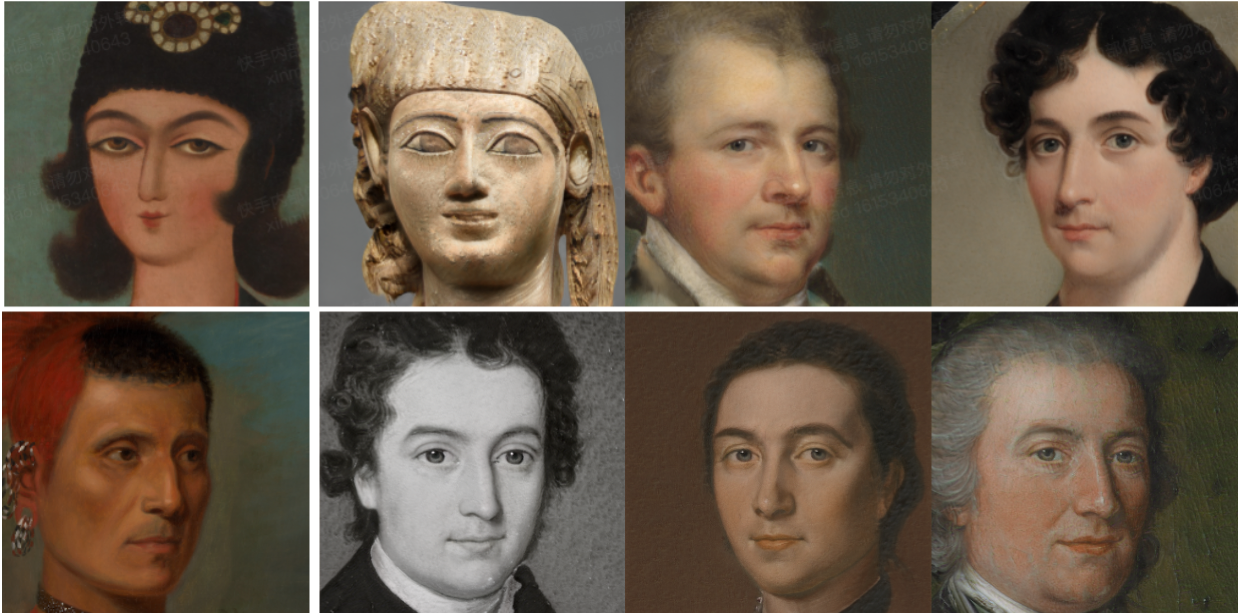}
\end{center}
\vspace{0.0cm}
\caption{Results for random vectors version StyleGAN on Metfaces. The selected training style images (left) with random assigned vectors can't provide useful artistic style information for output images (right).}
\label{fig:randomlabel}
\end{figure}

\begin{table}[t]
\begin{center}
\begin{tabular}{c|c}
\hline
Method & FID  \\
\hline
FGC StyleGAN with projection discriminator  &  {\bf 20.39}\\
\hline
FGC StyleGAN with concatenating discriminator & 25.34  \\
\hline
Vanilla StyleGAN &  29.73 \\
\hline
\end{tabular}
\end{center}
\caption{FID scores (lower is better) on Metafaces of FGC StyleGAN with projection discriminator versus concatenating discriminator, compared to vanilla StyleGAN.}
\label{tab:ablation_projection}
\end{table}

\begin{figure}[t]
\begin{center}
\includegraphics[width=1.\linewidth]{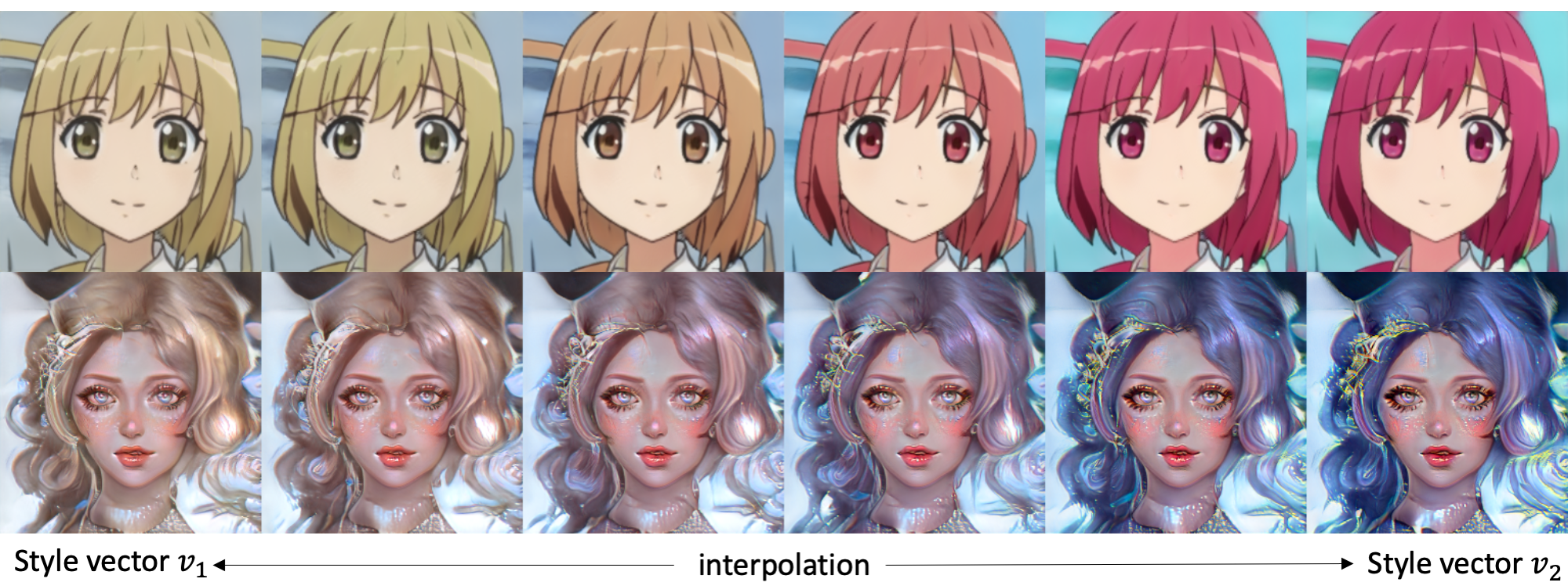}
\end{center}
\vspace{-.3cm}
\caption{New artistic styles (2nd to 5th column) can be generated by interpolating between 1st and 6th column.}
\label{fig:interpolation}
\end{figure}

\begin{figure}[t]
\begin{center}
\includegraphics[width=1.0\linewidth]{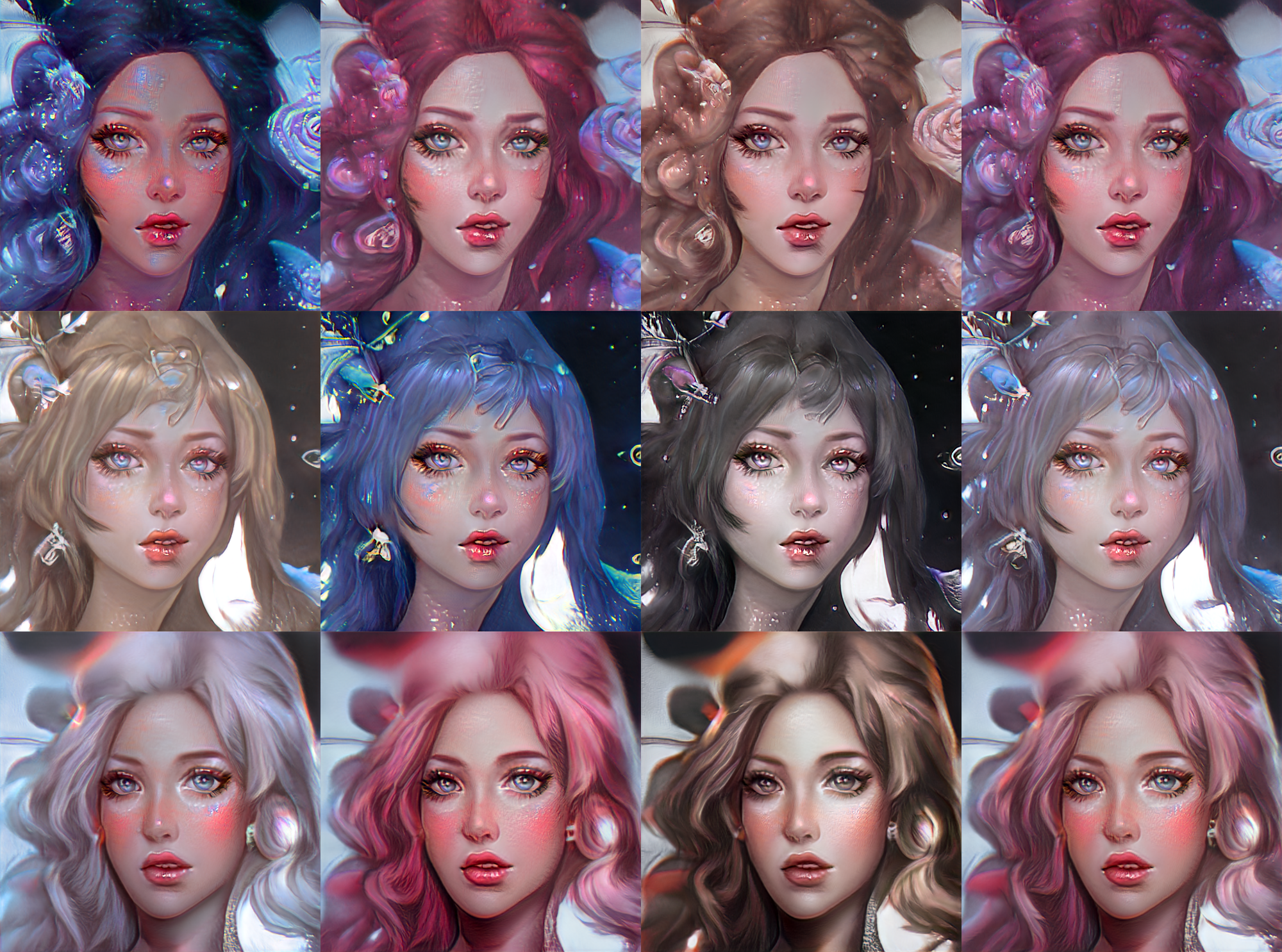}
\end{center}
\vspace{0.0cm}
\caption{New artistic styles (4th column) can be generated by interpolating between the style vectors corresponding to the 1st, 2nd and 3rd column.}
\label{fig:interpolation_3}
\end{figure}

\subsection{Ablation Study of the Embedded Style Space}
\label{sec:experiments:Fine-grained Space Analysis}
We first conduct an ablation study to verify the benefit of constructing the style space using training image embeddings. We create a control setup by assigning each image a fake style vector by randomly sample from Gaussian distribution with the same dimensionality as the style vector space. We train the random-style-vector version StyleGAN on Metfaces and compared the FID scores (Table~\ref{tab:randomlabel}). We can see that the FID score for FGC StyleGAN with random style space deteriorates to almost the same as vanilla StyleGAN. This has showed that the extracted fine-grained style vectors can help the network to better exploit the varying artistic style in training images, thereby better capture the data distribution in the learned generative model.

In Fig~\ref{fig:randomlabel} we show the images generated by the FGC StyleGAN with random style vectors. The first column of each row is the selected training image that provides a style vector $\mathbf{v}$, the other columns are the generated images from $\mathbf{v}$ (each row uses the same $\mathbf{v}$). We can see that this approach does not provide the artistic control as we have seen in FGC StyleGAN. This has further verified the importance of Gram matrix-based style space construction in our method.

\subsection{Ablation Study of the Projection Discriminator}
We then conduct an ablation study to verify the benefit of using the projection discriminator. We create a setup by substitute the projection discriminator with a  concatenating discriminator. Specifically, the style vector pass a fully connected layer first and then after reshaping we get a feature map has the same shape as the input image.The concatenating discriminator use the conditional information by concatenating that feature map to the input image. We compared the FID scores (Table~\ref{tab:ablation_projection}) by using different discriminators and the Vanilla version. We can see that compared to concatenating discriminator, the projection discriminator imporve the results by 4.95 in FID.

\subsection{Artistic Style Creation by Interpolation}
As mentioned before, our style space is continuous, so we can move in this space to generate different (and consistent) artistic styles that does not exist in the training set.

One easy way to verify this is by interpolating between existing artistic styles. Consider style vectors $\mathbf{v}_{1}$ and $\mathbf{v}_{2}$ corresponding to two training images. We generate a mixed style vector $\mathbf{v}_{3} = \lambda \cdot \mathbf{v}_{1} +(1-\lambda)\cdot \mathbf{v}_{2}$. Fig~\ref{fig:interpolation} shows the generated images by using this strategy with different $\lambda$ values $(0.8, \:0.6, \:0.4, \:0.2)$. The first and last column are generated images by style vector $\mathbf{v}_{1}$ and $\mathbf{v}_{2}$ respectively. Other columns are generated from $\mathbf{v}_{3}$. Even the generator has never seen the mixed style vector $\mathbf{v}_{3}$ during training, it still generates high quality images. The image generated from $\mathbf{v}_{3}$ reflects a natural mixture of artistic styles from $\mathbf{v}_{1}$ and $\mathbf{v}_{2}$. Similarly, as shown in Fig~\ref{fig:interpolation_3} we can mix three style vectors by $\mathbf{v}_{4} = \frac{1}{3} \mathbf{v}_{1} +\frac{1}{3} \mathbf{v}_{2}+\frac{1}{3}  \mathbf{v}_{3}$ and still generate high quality images.

%And the $v_{3}$ controlled output image also look like a mixed style images from two images on it's left in artistic style. 

\section{Conclusions}
In this paper, to  the  best  of  our  knowledge,  we  first  address  the fine-grained artistic style control problem. We presented an unsupervised approach to fine-grained control of artistic style in generating human portrays. Our method not only provides precise control of the style but also improves image quality as measured by FID. The continuous style space we constructed from image feature Gram matrix embedding shows desirable properties such as allowing the creation of new artistic styles by interpolating between existing ones. The ability to generate high quality images with precise style control from a very small number of training images (\emph{e.g.} 78 images from our game figure dataset) has great potential in online social entertainment applications such as real-time style transfer of selfies on a mobile phone.

{
\bibliography{egbib}

\begin{thebibliography}{33}
\providecommand{\natexlab}[1]{#1}

\bibitem[{Balakrishnan et~al.(2020)Balakrishnan, Xiong, Xia, and
  Perona}]{balakrishnan2020towards}
Balakrishnan, G.; Xiong, Y.; Xia, W.; and Perona, P. 2020.
\newblock Towards causal benchmarking of bias in face analysis algorithms.
\newblock In \emph{European Conference on Computer Vision}, 547--563. Springer.

\bibitem[{Brock, Donahue, and Simonyan(2018)}]{brock2018large}
Brock, A.; Donahue, J.; and Simonyan, K. 2018.
\newblock Large scale GAN training for high fidelity natural image synthesis.
\newblock \emph{arXiv preprint arXiv:1809.11096}.

\bibitem[{Chen et~al.(2016)Chen, Duan, Houthooft, Schulman, Sutskever, and
  Abbeel}]{chen2016infogan}
Chen, X.; Duan, Y.; Houthooft, R.; Schulman, J.; Sutskever, I.; and Abbeel, P.
  2016.
\newblock Infogan: Interpretable representation learning by information
  maximizing generative adversarial nets.
\newblock \emph{arXiv preprint arXiv:1606.03657}.

\bibitem[{Chen, Lai, and Liu(2018)}]{chen2018cartoongan}
Chen, Y.; Lai, Y.-K.; and Liu, Y.-J. 2018.
\newblock Cartoongan: Generative adversarial networks for photo cartoonization.
\newblock In \emph{Proceedings of the IEEE conference on computer vision and
  pattern recognition}, 9465--9474.

\bibitem[{Choi et~al.(2018)Choi, Choi, Kim, Ha, Kim, and
  Choo}]{choi2018stargan}
Choi, Y.; Choi, M.; Kim, M.; Ha, J.-W.; Kim, S.; and Choo, J. 2018.
\newblock Stargan: Unified generative adversarial networks for multi-domain
  image-to-image translation.
\newblock In \emph{Proceedings of the IEEE conference on computer vision and
  pattern recognition}, 8789--8797.

\bibitem[{Choi et~al.(2020)Choi, Uh, Yoo, and Ha}]{choi2020stargan}
Choi, Y.; Uh, Y.; Yoo, J.; and Ha, J.-W. 2020.
\newblock Stargan v2: Diverse image synthesis for multiple domains.
\newblock In \emph{Proceedings of the IEEE/CVF Conference on Computer Vision
  and Pattern Recognition}, 8188--8197.

\bibitem[{Deng et~al.(2009)Deng, Dong, Socher, Li, Li, and Fei}]{bb77703}
Deng, J.; Dong, W.; Socher, R.; Li, L.~J.; Li, K.; and Fei, L.~F. 2009.
\newblock ImageNet: {A} large-scale hierarchical image database.
\newblock In \emph{CVPR}, 248--255.

\bibitem[{Deng et~al.(2020)Deng, Yang, Chen, Wen, and
  Tong}]{deng2020disentangled}
Deng, Y.; Yang, J.; Chen, D.; Wen, F.; and Tong, X. 2020.
\newblock Disentangled and Controllable Face Image Generation via 3D
  Imitative-Contrastive Learning.
\newblock In \emph{Proceedings of the IEEE/CVF Conference on Computer Vision
  and Pattern Recognition}, 5154--5163.

\bibitem[{Gatys, Ecker, and Bethge(2015)}]{gatys2015texture}
Gatys, L.; Ecker, A.~S.; and Bethge, M. 2015.
\newblock Texture synthesis using convolutional neural networks.
\newblock \emph{Advances in neural information processing systems}, 28:
  262--270.

\bibitem[{Gatys, Ecker, and Bethge(2016)}]{gatys2016image}
Gatys, L.~A.; Ecker, A.~S.; and Bethge, M. 2016.
\newblock Image style transfer using convolutional neural networks.
\newblock In \emph{Proceedings of the IEEE conference on computer vision and
  pattern recognition}, 2414--2423.

\bibitem[{Goodfellow et~al.(2014)Goodfellow, Pouget-Abadie, Mirza, Xu,
  Warde-Farley, Ozair, Courville, and Bengio}]{NIPS2014_5ca3e9b1}
Goodfellow, I.; Pouget-Abadie, J.; Mirza, M.; Xu, B.; Warde-Farley, D.; Ozair,
  S.; Courville, A.; and Bengio, Y. 2014.
\newblock Generative Adversarial Nets.
\newblock In Ghahramani, Z.; Welling, M.; Cortes, C.; Lawrence, N.; and
  Weinberger, K.~Q., eds., \emph{Advances in Neural Information Processing
  Systems}, volume~27, 2672--2680. Curran Associates, Inc.

\bibitem[{H{\"a}rk{\"o}nen et~al.(2020)H{\"a}rk{\"o}nen, Hertzmann, Lehtinen,
  and Paris}]{harkonen2020ganspace}
H{\"a}rk{\"o}nen, E.; Hertzmann, A.; Lehtinen, J.; and Paris, S. 2020.
\newblock Ganspace: Discovering interpretable gan controls.
\newblock \emph{arXiv preprint arXiv:2004.02546}.

\bibitem[{Huang and Belongie(2017)}]{huang2017arbitrary}
Huang, X.; and Belongie, S. 2017.
\newblock Arbitrary style transfer in real-time with adaptive instance
  normalization.
\newblock In \emph{Proceedings of the IEEE International Conference on Computer
  Vision}, 1501--1510.

\bibitem[{Isola et~al.(2017)Isola, Zhu, Zhou, and Efros}]{isola2017image}
Isola, P.; Zhu, J.-Y.; Zhou, T.; and Efros, A.~A. 2017.
\newblock Image-to-image translation with conditional adversarial networks.
\newblock In \emph{Proceedings of the IEEE conference on computer vision and
  pattern recognition}, 1125--1134.

\bibitem[{Jahanian, Chai, and Isola(2019)}]{jahanian2019steerability}
Jahanian, A.; Chai, L.; and Isola, P. 2019.
\newblock On the" steerability" of generative adversarial networks.
\newblock \emph{arXiv preprint arXiv:1907.07171}.

\bibitem[{Karras et~al.(2017)Karras, Aila, Laine, and
  Lehtinen}]{karras2017progressive}
Karras, T.; Aila, T.; Laine, S.; and Lehtinen, J. 2017.
\newblock Progressive growing of gans for improved quality, stability, and
  variation.
\newblock \emph{arXiv preprint arXiv:1710.10196}.

\bibitem[{Karras et~al.(2020{\natexlab{a}})Karras, Aittala, Hellsten, Laine,
  Lehtinen, and Aila}]{karras2020training}
Karras, T.; Aittala, M.; Hellsten, J.; Laine, S.; Lehtinen, J.; and Aila, T.
  2020{\natexlab{a}}.
\newblock Training generative adversarial networks with limited data.
\newblock \emph{arXiv preprint arXiv:2006.06676}.

\bibitem[{Karras, Laine, and Aila(2019)}]{karras2019style}
Karras, T.; Laine, S.; and Aila, T. 2019.
\newblock A style-based generator architecture for generative adversarial
  networks.
\newblock In \emph{Proceedings of the IEEE conference on computer vision and
  pattern recognition}, 4401--4410.

\bibitem[{Karras et~al.(2020{\natexlab{b}})Karras, Laine, Aittala, Hellsten,
  Lehtinen, and Aila}]{karras2020analyzing}
Karras, T.; Laine, S.; Aittala, M.; Hellsten, J.; Lehtinen, J.; and Aila, T.
  2020{\natexlab{b}}.
\newblock Analyzing and improving the image quality of stylegan.
\newblock In \emph{Proceedings of the IEEE/CVF Conference on Computer Vision
  and Pattern Recognition}, 8110--8119.

\bibitem[{Kim et~al.(2019)Kim, Kim, Kang, and Lee}]{kim2019u}
Kim, J.; Kim, M.; Kang, H.; and Lee, K. 2019.
\newblock U-GAT-IT: unsupervised generative attentional networks with adaptive
  layer-instance normalization for image-to-image translation.
\newblock \emph{arXiv preprint arXiv:1907.10830}.

\bibitem[{Kingma and Welling(2014)}]{journals/corr/KingmaW13}
Kingma, D.~P.; and Welling, M. 2014.
\newblock Auto-Encoding Variational Bayes.
\newblock In Bengio, Y.; and LeCun, Y., eds., \emph{2nd International
  Conference on Learning Representations, ICLR 2014, Banff, AB, Canada, April
  14-16, 2014, Conference Track Proceedings}.

\bibitem[{Kowalski et~al.(2020)Kowalski, Garbin, Estellers, Baltru{\v{s}}aitis,
  Johnson, and Shotton}]{kowalski2020config}
Kowalski, M.; Garbin, S.~J.; Estellers, V.; Baltru{\v{s}}aitis, T.; Johnson,
  M.; and Shotton, J. 2020.
\newblock CONFIG: Controllable Neural Face Image Generation.
\newblock \emph{arXiv preprint arXiv:2005.02671}.

\bibitem[{Lee et~al.(2009)Lee, Grosse, Ranganath, and Ng}]{conf/icml/LeeGRN09}
Lee, H.; Grosse, R.~B.; Ranganath, R.; and Ng, A.~Y. 2009.
\newblock Convolutional deep belief networks for scalable unsupervised learning
  of hierarchical representations.
\newblock In Danyluk, A.~P.; Bottou, L.; and Littman, M.~L., eds.,
  \emph{Proceedings of the 26th Annual International Conference on Machine
  Learning, ICML 2009, Montreal, Quebec, Canada, June 14-18, 2009}, volume 382
  of \emph{ACM International Conference Proceeding Series}, 609--616. ACM.
\newblock ISBN 978-1-60558-516-1.

\bibitem[{Liu, Breuel, and Kautz(2017)}]{liu2017unsupervised}
Liu, M.-Y.; Breuel, T.; and Kautz, J. 2017.
\newblock Unsupervised image-to-image translation networks.
\newblock \emph{arXiv preprint arXiv:1703.00848}.

\bibitem[{Miyato et~al.(2018)Miyato, Kataoka, Koyama, and
  Yoshida}]{miyato2018spectral}
Miyato, T.; Kataoka, T.; Koyama, M.; and Yoshida, Y. 2018.
\newblock Spectral normalization for generative adversarial networks.
\newblock \emph{arXiv preprint arXiv:1802.05957}.

\bibitem[{Miyato and Koyama(2018)}]{miyato2018cgans}
Miyato, T.; and Koyama, M. 2018.
\newblock cGANs with projection discriminator.
\newblock \emph{arXiv preprint arXiv:1802.05637}.

\bibitem[{Shen et~al.(2020)Shen, Gu, Tang, and Zhou}]{shen2020interpreting}
Shen, Y.; Gu, J.; Tang, X.; and Zhou, B. 2020.
\newblock Interpreting the latent space of gans for semantic face editing.
\newblock In \emph{Proceedings of the IEEE/CVF Conference on Computer Vision
  and Pattern Recognition}, 9243--9252.

\bibitem[{Shoshan et~al.(2021)Shoshan, Bhonker, Kviatkovsky, and
  Medioni}]{shoshan2021gan}
Shoshan, A.; Bhonker, N.; Kviatkovsky, I.; and Medioni, G. 2021.
\newblock GAN-Control: Explicitly Controllable GANs.
\newblock \emph{arXiv preprint arXiv:2101.02477}.

\bibitem[{Tenenbaum and Freeman(1997)}]{nips-9:Tenenbaum+Freeman:1997}
Tenenbaum, J.~B.; and Freeman, W.~T. 1997.
\newblock Separating Style and Content.
\newblock In Mozer, M.~C.; Jordan, M.~I.; and Petsche, T., eds., \emph{Advances
  in Neural Information Processing Systems}, volume~9, 662. The {MIT} Press.

\bibitem[{Tewari et~al.(2020)Tewari, Elgharib, Bharaj, Bernard, Seidel,
  P{\'e}rez, Zollhofer, and Theobalt}]{tewari2020stylerig}
Tewari, A.; Elgharib, M.; Bharaj, G.; Bernard, F.; Seidel, H.-P.; P{\'e}rez,
  P.; Zollhofer, M.; and Theobalt, C. 2020.
\newblock Stylerig: Rigging stylegan for 3d control over portrait images.
\newblock In \emph{Proceedings of the IEEE/CVF Conference on Computer Vision
  and Pattern Recognition}, 6142--6151.

\bibitem[{Wang et~al.(2018)Wang, Liu, Zhu, Tao, Kautz, and
  Catanzaro}]{wang2018high}
Wang, T.-C.; Liu, M.-Y.; Zhu, J.-Y.; Tao, A.; Kautz, J.; and Catanzaro, B.
  2018.
\newblock High-resolution image synthesis and semantic manipulation with
  conditional gans.
\newblock In \emph{Proceedings of the IEEE conference on computer vision and
  pattern recognition}, 8798--8807.

\bibitem[{Yang, Shen, and Zhou(2021)}]{yang2021semantic}
Yang, C.; Shen, Y.; and Zhou, B. 2021.
\newblock Semantic hierarchy emerges in deep generative representations for
  scene synthesis.
\newblock \emph{International Journal of Computer Vision}, 1--16.

\bibitem[{Zhu et~al.(2017)Zhu, Park, Isola, and Efros}]{zhu2017unpaired}
Zhu, J.-Y.; Park, T.; Isola, P.; and Efros, A.~A. 2017.
\newblock Unpaired image-to-image translation using cycle-consistent
  adversarial networks.
\newblock In \emph{Proceedings of the IEEE international conference on computer
  vision}, 2223--2232.

\end{thebibliography}
}

\end{document}